\documentclass[letterpaper, 10 pt, conference]{ieeeconf}  

\usepackage{times}
\usepackage[numbers]{natbib}
\usepackage{multicol}
\usepackage[bookmarks=true]{hyperref}

\usepackage{lipsum}
\usepackage{comment}
\usepackage{balance}
\usepackage{graphicx}
\usepackage[table]{xcolor}
\usepackage{caption}
\usepackage{listings}
\usepackage{xcolor}

\definecolor{codegreen}{rgb}{0,0.6,0}
\definecolor{codegray}{rgb}{0.5,0.5,0.5}
\definecolor{codepurple}{rgb}{0.58,0,0.82}
\definecolor{backcolour}{rgb}{0.95,0.95,0.92}

\lstdefinestyle{mystyle}{
    backgroundcolor=\color{backcolour},
    commentstyle=\color{codegreen},
    keywordstyle=\color{magenta},
    numberstyle=\tiny\color{codegray},
    stringstyle=\color{codepurple},
    basicstyle=\ttfamily,
    breaklines=true,
    captionpos=b,
    keepspaces=true,
    numbers=left,
    numbersep=5pt,
    showstringspaces=false,
    tabsize=2
}
\usepackage{minted}
\usepackage{xcolor}
\definecolor{bg}{HTML}{f2f2ea}
\usepackage{float} 

\usepackage{seqsplit}

\usepackage{amssymb}
\usepackage{amsmath}
\usepackage{booktabs}
 
\usepackage{graphicx} 
\usepackage{subcaption}
\usepackage{bbm}

\IEEEoverridecommandlockouts                              

\usepackage{tabularx} 

\title{\Large \bf
Hybrid Diffusion Policies with Projective Geometric Algebra for Efficient Robot Manipulation Learning
}

\author{Xiatao Sun$^{1}$, Yuxuan Wang$^{2}$, Shuo Yang$^{3}$, Yinxing Chen$^{1}$, Daniel Rakita$^{1}$
\thanks{$^{1}$Xiatao Sun, Yinxing Chen, Daniel Rakita are with the Department of Computer Science, Yale University,
        New Haven, CT 06520, USA
        {\tt\small \{xiatao.sun, j.y.chen, daniel.rakita\}@yale.edu}}
\thanks{$^{2}$Yuxuan Wang is with the Department of Computer and Information Science, University of Pennsylvania,
        Philadelphia, PA 19104, USA
        {\tt\small wangy49@seas.upenn.edu}}
\thanks{$^{3}$Shuo Yang is with the Department of Electrical and Systems Engineering, University of Pennsylvania,
        Philadelphia, PA 19104, USA
        {\tt\small  yangs1@seas.upenn.edu}}
\thanks{This work was supported by Office of Naval Research award N00014-24-1-2124}
}    

\begin{document}

\maketitle
\thispagestyle{empty}
\pagestyle{empty}


\begin{abstract}
Diffusion policies are a powerful paradigm for robot learning, but their training is often inefficient. A key reason is that networks must relearn fundamental spatial concepts, such as translations and rotations, from scratch for every new task. To alleviate this redundancy, we propose embedding geometric inductive biases directly into the network architecture using Projective Geometric Algebra (PGA). PGA provides a unified algebraic framework for representing geometric primitives and transformations, allowing neural networks to reason about spatial structure more effectively. In this paper, we introduce hPGA-DP, a novel hybrid diffusion policy that capitalizes on these benefits. Our architecture leverages the Projective Geometric Algebra Transformer (P-GATr) as a state encoder and action decoder, while employing established U-Net or Transformer-based modules for the core denoising process. Through extensive experiments and ablation studies in both simulated and real-world environments, we demonstrate that hPGA-DP significantly improves task performance and training efficiency. Notably, our hybrid approach achieves substantially faster convergence compared to both standard diffusion policies and architectures that rely solely on P-GATr. The project website is available at: \href{https://apollo-lab-yale.github.io/26-ICRA-hPGA-website/}{https://apollo-lab-yale.github.io/26-ICRA-hPGA-website/}.
\end{abstract}
\section{Introduction}
\label{sec:introduction}

Diffusion policies \cite{chi2023diffusion} have emerged as a powerful paradigm for visuomotor control in robotics, offering reliable convergence through iterative denoising of action trajectories. These models are typically trained from scratch with hundreds of epochs, and while effective, this training paradigm presents a critical inefficiency: the network must repeatedly relearn basic spatial priors---such as translations and rotations---for every new task or environment. This redundant relearning not only inflates the computational cost but also slows down convergence. 

Given that spatial concepts are inherently universal across robotic tasks, integrating geometric inductive biases directly into the network architecture presents a compelling strategy to alleviate this redundancy. Projective Geometric Algebra (PGA), a mathematical framework offering a unified representation of spatial entities and operations through mathematical objects called \textit{multivectors}, is particularly suited to embedding such biases. Multivectors provide a structured algebraic and geometric representation for spatial priors like points, translations, and rotations, which enables neural networks to perform spatial computations more efficiently and intuitively, thereby potentially improving training efficiency and task performance. For a thorough and practical treatment of PGA, we refer the reader to the work by \citet{dorst2022guided}.

To incorporate PGA into general learning tasks, prior research has proposed the Projective Geometric Algebra Transformer (P-GATr) \cite{brehmer2023geometric}, demonstrating its effectiveness in a set of initial spatial learning tasks, such as simulated n-body modeling and predicting shear stress in human arteries, outperforming non-geometric architectures. However, incorporating P-GATr directly as a denoising backbone within diffusion policies is challenging in robotics contexts. As our experiments suggest, the inherent geometric inductive biases and the complexity of multivector computations within P-GATr make it difficult to directly learn noise prediction for effective denoising, resulting in prohibitively slow convergence.

To overcome these limitations, we propose the \textbf{h}ybrid \textbf{P}rojective \textbf{G}eometric \textbf{A}lgebra \textbf{D}iffusion \textbf{P}olicy (hPGA-DP), a hybrid diffusion policy architecture that predicts an action sequence based on an observation sequence of the concatenations of the proprioceptive states of the robot and poses of task-relevant objects. The approach is designed to leverage the strengths of both geometric and traditional neural network approaches. Specifically, we employ P-GATr as a spatial state encoder and action decoder, while utilizing established architectures such as U-Net \cite{ronneberger2015u} or Transformer \cite{vaswani2017attention} for the denoising module. This hybrid approach allows the strong geometric inductive biases of P-GATr to efficiently embed spatial structures into a representation space amenable to effective denoising, while simultaneously benefiting from the proven denoising capabilities of conventional architectures. To our knowledge, our work is the first to incorporate PGA into network architecture for diffusion policies.

We evaluate and validate hPGA-DP through several experiments and ablation studies in both simulated and real-world scenarios. Our results demonstrate that hPGA-DP successfully addresses the convergence issues observed when solely relying on P-GATr, achieving significantly faster convergence and superior task performance compared to standard Transformer or U-Net-based diffusion policy architectures. 

We open-source our code to facilitate future research on diffusion policies with geometric algebra.\footnote{\href{https://github.com/Apollo-Lab-Yale/hpga-dp}{https://github.com/Apollo-Lab-Yale/hpga-dp}}

\section{Related Works}
\label{sec:related_works}

\subsection{Diffusion Policy}

Diffusion models, originally proposed by \citet{ho2020denoising}, generate data by reversing a stochastic forward process, and have achieved remarkable success in image and video synthesis \cite{ podell2023sdxl, guo2023animatediff}. These models have since been adapted for robotic motion generation, with diffusion policies becoming a prominent paradigm for robot learning \cite{chi2023diffusion, wang2025hierarchical}, albeit requiring extensive training over hundreds of epochs.

To mitigate this issue, several works aim to improve training or inference efficiency: \citet{ze20243d} proposed using point clouds for richer input; \citet{wang2024equivariant} imposed symmetry in denoising to boost generalization; \citet{sun2025dynamic} exploited low-rank overparameterization to accelerate training; and \citet{reuss2024efficient} employed mixture-of-experts to reduce inference cost.

While effective, these efforts largely target data efficiency, modality, or inference, without altering the learnable network backbone. Prior architectural innovations centered on Transformers \cite{hou2024diffusion, wang2025mtdp}, yielding only modest improvements. In contrast, our hPGA-DP introduces a novel hybrid architecture that achieves substantially better training efficiency and policy performance.

\subsection{Geometric Algebra for Robot Learning} 

Geometric Algebra (GA) offers a unified framework for representing geometric entities and transformations \cite{dorst2009geometric}. Among its variants, Projective Geometric Algebra (PGA) is tailored for Euclidean geometry, using planes as primitives to represent motions compactly with just four basis elements \cite{dorst2022guided, roelfs2023graded}. While GA is mathematically mature, its adoption in robotics and machine learning remains limited, with most applications favoring Conformal Geometric Algebra (CGA) \cite{zhu2025conformal, carbajal2024fika} due to its symbolic compactness, as seen in manipulation control \cite{low2023geometric} and tactile ergodic control \cite{bilaloglu2024tactile}.

For robotics tasks grounded in Euclidean spaces, PGA ($\mathbb{G}_{3,0,1}$) offers a computationally simpler yet expressive alternative \cite{brehmer2023geometric, ruhe2023clifford}. Existing PGA applications have largely focused on dynamics modeling \cite{sun2023analytical, sun2023high}, with limited exploration in learning due to the scarcity of GA-compatible neural architectures \cite{liu2022geometric, zhong2023geometric}. The recent Projective Geometric Algebra Transformer (P-GATr) \cite{brehmer2023geometric} addresses this gap by embedding geometric inductive biases for geometric learning \cite{brehmer2024lorentz, lee2023towards}. 
Unlike the original P-GATr work, which applied a model-based diffuser via reinforcement learning \cite{janner2022planning} only in numerical simulations, we introduce the first integration of P-GATr into a diffusion policy via imitation learning \cite{chi2023diffusion}, validating it on both complex physical simulations and real-world tasks.

In robotics, a concurrent work \cite{zhong2025gagrasp} applied P-GATr within diffusion models for grasp generation, but the approach is limited to static tasks and requires long training times. In contrast, our proposed hPGA-DP is an end-to-end diffusion policy architecture that achieves superior performance and faster convergence compared to baseline methods across a broad range of robotic learning tasks.

\section{Technical Overview}
\label{sec:technical_overview}

\begin{figure*}[t]
\centering
\includegraphics[width=\textwidth]
{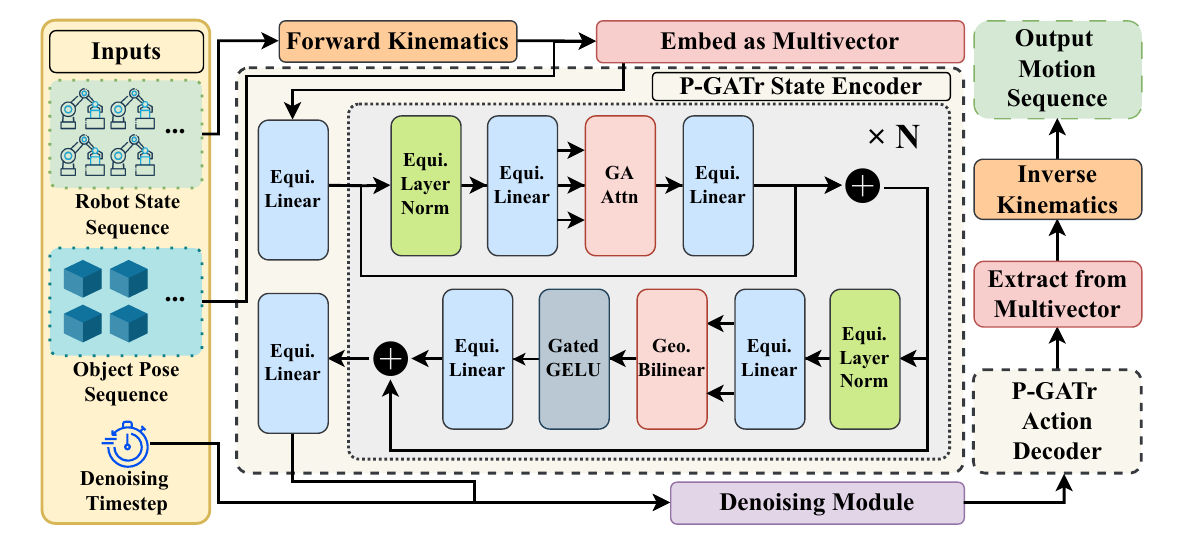}
\caption{Overview of the hPGA-DP network architecture. }
\label{fig:architecture}
\vspace{-15pt}
\end{figure*}

In this section, we present an overview of our proposed hPGA-DP approach.  The goal of hPGA-DP is to learn a receding horizon policy $\pi_\theta$ that outputs a sequence of actions $\mathbf{a}^{t:t+H_p}$ conditioned on a sequence of observations $\mathbf{o}^{t-H_o:t}$, where $H_p$ and $H_o$ denote the action prediction and observation horizons, respectively.

\subsection{Observations}  
Each observation at time \( t \) is defined as  $\mathbf{o}_t = \left[ \mathbf{s}_t, \left\{ \mathbf{T}_i^t \right\}_{i=1}^{J} \right],$
where \( \mathbf{s}_t \in \mathbb{R}^{d_s} \) represents the robot state, and \( \left\{ \mathbf{T}_i^t \right\}_{i=1}^J \) denotes the spatial poses of the \( J \) task-relevant objects. Each pose $ \mathbf{T}_i^t $ consists of a 3D position vector and a unit quaternion representing orientation. 

\subsection{Actions} Each action at time $\mathbf{a}_t$ may include spatial positions and/ or orientations for key links on a robot or any other scalar properties that can affect change on the robot platform.  For instance, an action may include the end-effector position at the given time, $\mathbf{p}_{ee}^t \in \mathbb{R}^{3}$, the end-effector orientation at the given time, $\mathbf{q}_{ee}^t \in \mathbb{H}_1$ (the space of unit-quaternions), and scalar representing how open or closed the gripper is at the given time $g^t \in \mathbb{R}$.  The action representation is flexible and can include any number platform-specific properties, accommodating various embodiments such as bimanual or humanoid robots with multiple end-effectors.  We assume that any system adopting our approach can convert these high-level actions into corresponding robot states or control commands, such as through an inverse kinematics solver.

\subsection{Architecture} The network architecture for hPGA-DP is illustrated in Fig.~\ref{fig:architecture}. First, the robot states from the input observation sequence are converted into the positions and orientations of key links, such as end-effectors, using the robot's forward kinematics model, in a manner similar to the action representation described above. These spatial components, combined with the poses of task-relevant objects, are then transformed into \textit{multivectors}, the central representational objects in geometric algebra. For the detailed conversions to multivectors, we refer the reader to the practical treatment by \citet{dorst2022guided}.

The resulting multivectors are stacked in a tensor $\mathbf{x}_o^{t-H_o:t} \in \mathbb{R}^{H_o \times K_o \times 16}$, where $H_o$ is the observation horizon, $K_o$ is the total number positions and orientations of key links on the robot and task-relevant objects in the observations, and $16$ is the length of a multivector in $\mathbb{G}_{3,0,1}$.  The multivector stack is then processed by a P-GATr state encoder \citep{brehmer2023geometric} to produce an observation latent, $\mathbf{z}_{\mathbf{o}} \in \mathbb{R}^{H_o \times K_o \times 16}$.  Note that this latent representation maintains the same dimensionality as the input multivector stack.

The observation latent tensor $\mathbf{z}_o$ is passed to a denoising module, implemented as either a Transformer or U-Net, which produces denoised action latents \( \mathbf{z}_{\mathbf{a}} \in \mathbb{R}^{H_p \times K_a \times 16} \). Here, \( H_p \) denotes the prediction horizon, \( K_a \) is the total number of positional, rotational, or scalar components in an action, and 16 is the dimensionality of a multivector in $\mathbb{G}_{3,0,1}$.  These action latents are then decoded by a P-GATr action decoder \citep{brehmer2023geometric}, which mirrors the structure of the state encoder. 

The decoder produces a stack of action multivectors as a tensor \( \mathbf{x}_a^{t:t+H_p} \in \mathbb{R}^{H_p \times K_a \times 16} \), where each multivector represents a predicted position, orientation, or scalar action component. This tensor is unpacked into individual multivectors corresponding to specific link or scalar properties at each time step, and then converted into standard geometric representations such as 3D positions, unit quaternions, and scalar values (e.g., for gripper control). The resulting outputs can be passed to a controller or inverse kinematics solver to recover the final sequence of robot states.

A more in-depth explanation and design justification for this architecture is presented in \S\ref{sec:technical_details}.

\subsection{Projective Geometric Algebra Transformer (P-GATr)}  
For both the state encoder and action decoder in our architecture, we employ the P-GATr model and associated network primitives introduced by \citet{brehmer2023geometric}. Structurally, P-GATr resembles a standard Transformer \citep{vaswani2017attention} with pre-layer normalization \citep{xiong2020layer, baevskiadaptive}. Each of the \( N \) Transformer blocks in P-GATr includes an Equivariant Linear layer (Equi. Linear), Geometric Bilinear layers (Geo. Bilinear), Multivector Attention (GA Attn), a Scalar-Gated Gaussian Error Linear Unit (Gated GELU) \citep{hendrycks2016gaussian}, and an \( \mathrm{E}(3) \)-Equivariant LayerNorm (Equi. LayerNorm). A detailed description of these components can be found in the original P-GATr paper by \citet{brehmer2023geometric}.

\section{Architecture Design Choices and Details}
\label{sec:technical_details}

In this section, we describe the network design choices and training procedure of hPGA-DP.  In preliminary experiments, we initially employed P-GATr directly as the denoising network, aiming to leverage its strong geometric inductive bias. However, we observed that this naive integration resulted in impractically slow convergence, consistent with results in concurrent work by \citet{zhong2025gagrasp}, which reports week-long training times. We hypothesize that this inefficiency stems from a fundamental mismatch in inductive priors: P-GATr is tailored for processing structured geometric data, whereas the objective of the denoising module is to reverse a stochastic process.

Based on this observation, we design hPGA-DP such that traditional architectures, such as Transformers or U-Nets, serve as the denoising backbone, while P-GATr is utilized as an encoder for observations and a decoder for actions. This allows the denoising process to operate in a latent space where geometric structure is learned implicitly through P-GATr, but without constraining the denoising network itself to a deterministic geometric bias.

Following standard diffusion frameworks~\cite{chi2023diffusion, ho2020denoising}, we apply the forward noising process to the action latent space:
\begin{equation*}
\mathbf{z}_{\mathbf{a}, k} = \sqrt{\bar{\alpha}_k} \, \mathbf{z}_{\mathbf{a}, 0} + \sqrt{1 - \bar{\alpha}_k} \, \boldsymbol{\varepsilon}, \quad \boldsymbol{\varepsilon} \sim \mathcal{N}(0, \mathbf{I}),
\end{equation*}
where $\mathbf{z}_{\mathbf{a}, k}$ denotes the noisy latent at denoising step $k$, $\mathbf{z}_{\mathbf{a}, 0}$ is the clean latent representation of the action sequence, $\bar{\alpha}_k$ is the cumulative signal retention factor from the noise scheduler that monotonically decreases as $k$ increases, and $\boldsymbol{\varepsilon}$ is sampled from a standard Gaussian. 

The state encoder and denoising module $\epsilon_\theta$ are trained to predict the added noise, with the following mean squared error loss:
$$
\mathcal{L}_{\text{Encode\&Denoise}} =
\left\| \epsilon_\theta(\mathbf{z}_{\mathbf{a}, k}, \mathbf{z}_{\mathbf{o}}, k) - \boldsymbol{\varepsilon} \right\|^2.
$$
where $\epsilon_\theta$ takes as input the noisy latent $\mathbf{z}_{\mathbf{a}, k}$ at timestep $k$, and is conditioned on the observation latent $\mathbf{z}_{\mathbf{o}}$ to predict the noise $\boldsymbol{\varepsilon}$.

\begin{figure*}[t]
\centering
\begin{minipage}{\textwidth}
    \begin{subfigure}{\textwidth}
        \centering
        \begin{subfigure}[b]{0.195\textwidth}
            \centering
            \includegraphics[width=\linewidth]{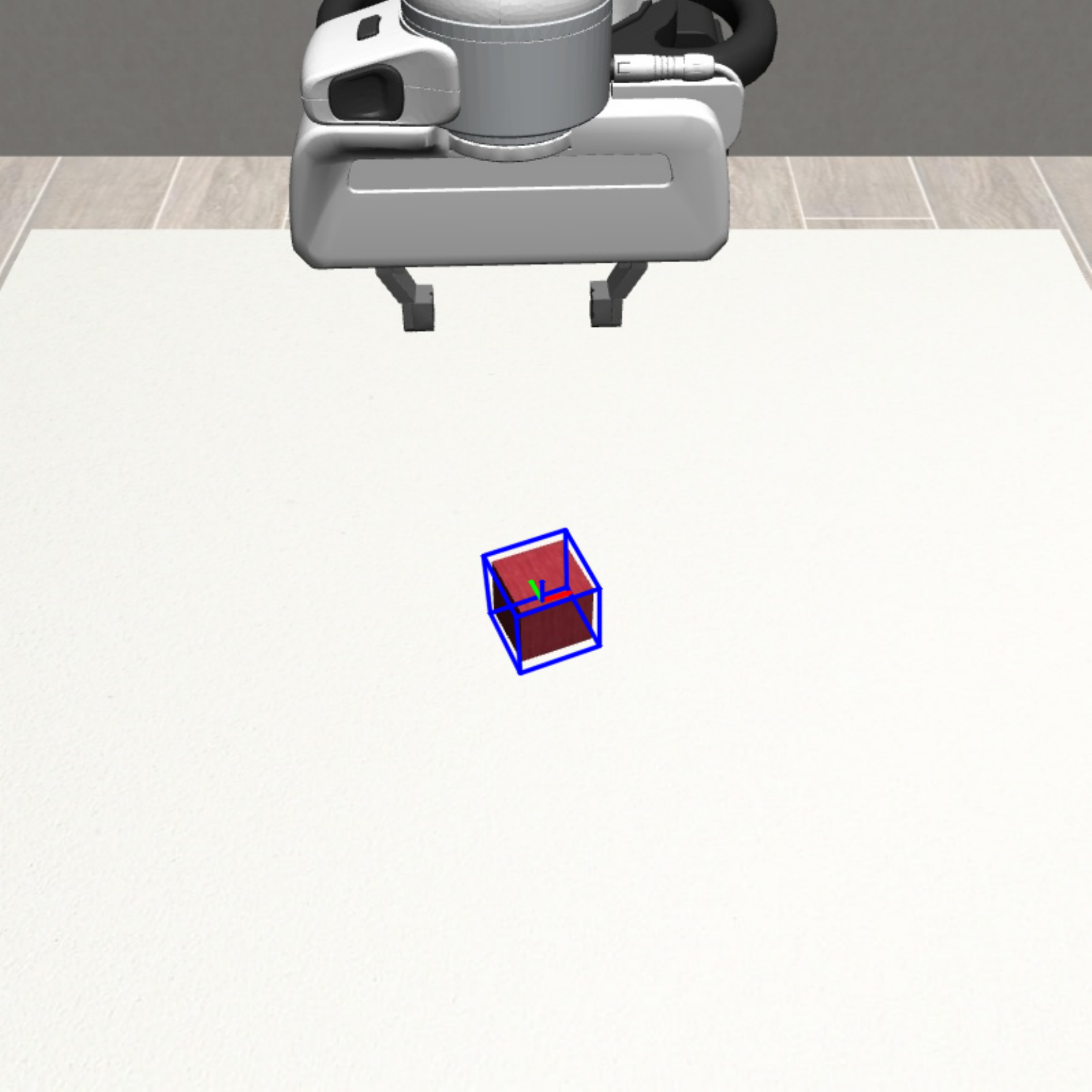}
            \caption*{Lift}
        \end{subfigure}
        \hfill
        \begin{subfigure}[b]{0.195\textwidth}
            \centering
            \includegraphics[width=\linewidth]{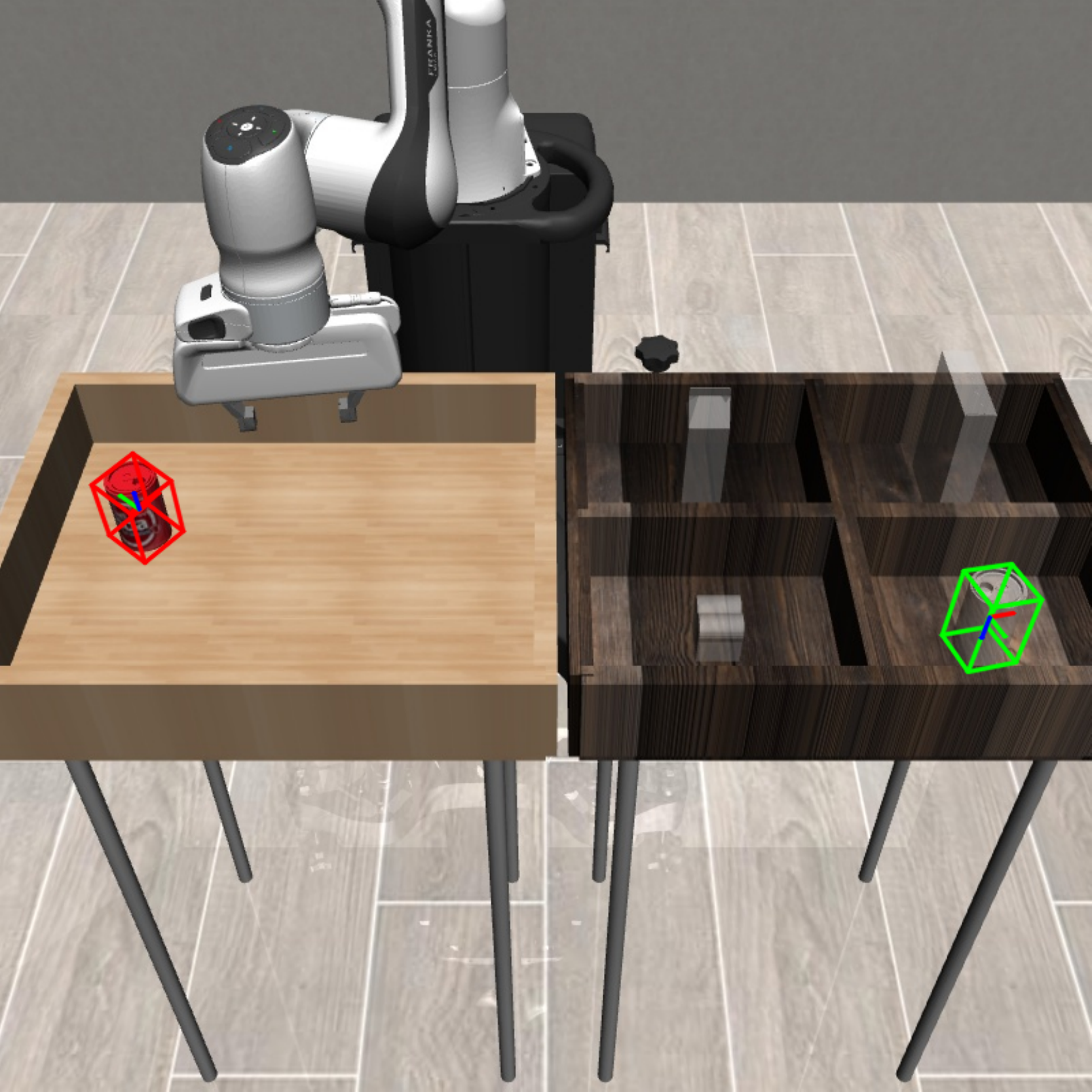}
            \caption*{Can}
        \end{subfigure}
        \hfill
        \begin{subfigure}[b]{0.195\textwidth}
            \centering
            \includegraphics[width=\linewidth]{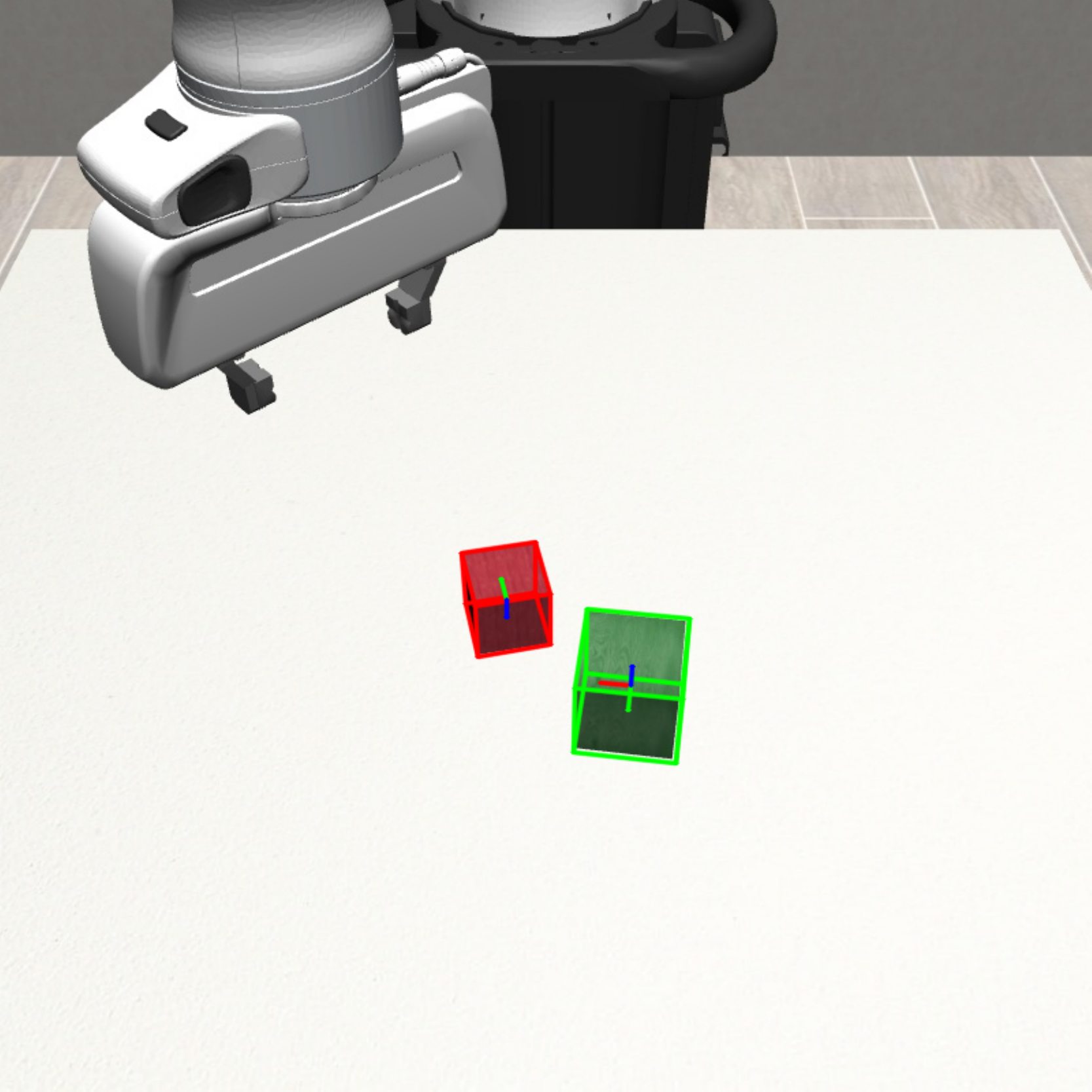}
            \caption*{Stack}
        \end{subfigure}
        \hfill
        \begin{subfigure}[b]{0.195\textwidth}
            \centering
            \includegraphics[width=\linewidth]{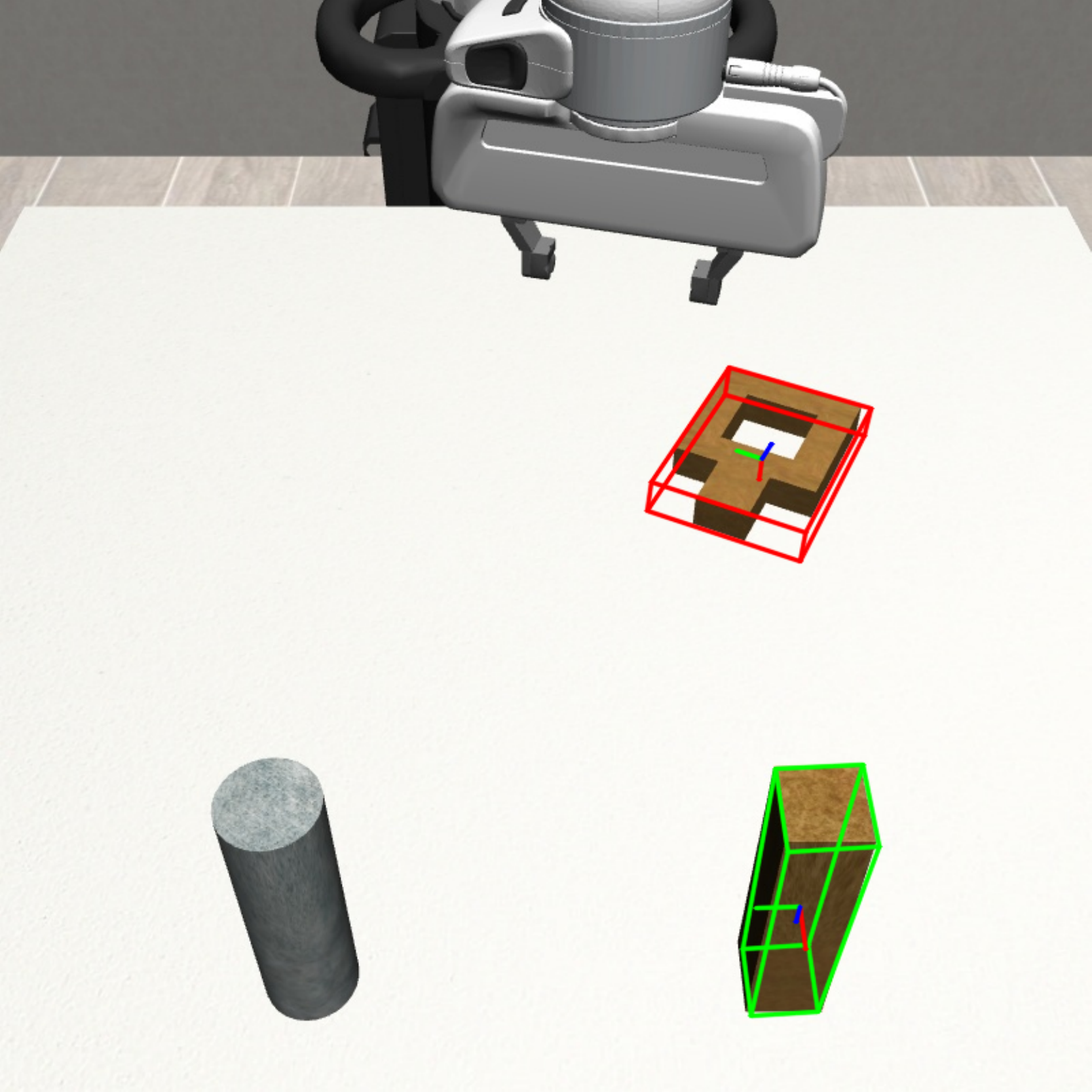}
            \caption*{Square}
        \end{subfigure}
        \hfill
        \begin{subfigure}[b]{0.195\textwidth}
            \centering
            \includegraphics[width=\linewidth]{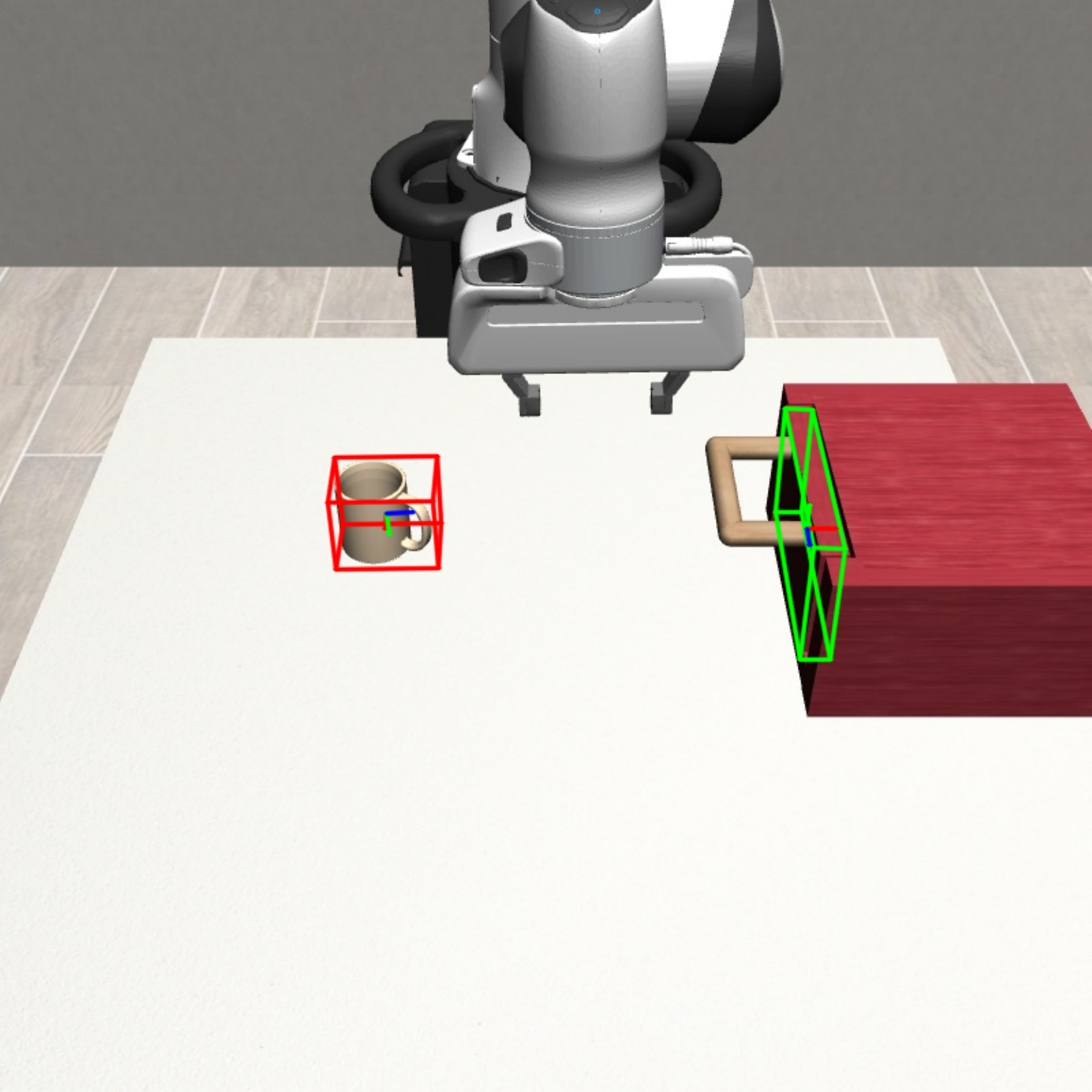}
            \caption*{Mug}
        \end{subfigure}
    \end{subfigure}

    \vspace{0.5cm}

    \begin{minipage}{0.48\textwidth}
        \centering
        \begin{tabular}{l *{5}{c}|c}
            \toprule
            \textbf{Network} & \textbf{Lift} & \textbf{Can} & \textbf{Stack} & \textbf{Square} & \textbf{Mug} & \textbf{MET} \\
            \midrule
            U-Net (State) & $0.54$ & $0.48$ & $0.22$ & $0.50$ & $0.48$ & $3.12$ \\
            Trans. (State) & $0.18$ & $0.42$ & $0.62$ & $0.56$ & $0.32$ & $3.46$ \\
            P-GATr (State) & $0.00$ & $0.00$ & $0.00$ & $0.00$ & $0.00$ & $23.49$ \\
            U-Net (6D) & $0.48$ & $0.42$ & $0.20$ & $0.46$ & $0.42$ & $\mathbf{3.08}$ \\
            Trans. (6D) & $0.12$ & $0.44$ & $0.60$ & $0.54$ & $0.30$ & $3.49$ \\
            P-GATr (6D) & $0.00$ & $0.00$ & $0.00$ & $0.00$ & $0.00$ & $23.75$ \\
            \midrule
            hPGA-U (State) & $0.98$ & $0.96$ & $\mathbf{0.98}$ & $\mathbf{0.84}$ & $\mathbf{0.72}$ & $4.35$\\
            hPGA-T (State) & $\mathbf{1.00}$ & $\mathbf{0.98}$ & $0.82$ & $0.70$ & $0.62$ & $4.61$ \\
            hPGA-U (6D) & $0.96$ & $0.90$ & $0.96$ & $0.78$ & $0.68$ & $4.42$\\
            hPGA-T (6D) & $0.98$ & $0.92$ & $0.84$ & $0.68$ & $0.60$ & $4.66$\\
            \bottomrule
        \end{tabular}
    \end{minipage}
    \hfill
    \begin{minipage}{0.6\textwidth}
        \centering
        \includegraphics[width=0.7\linewidth]{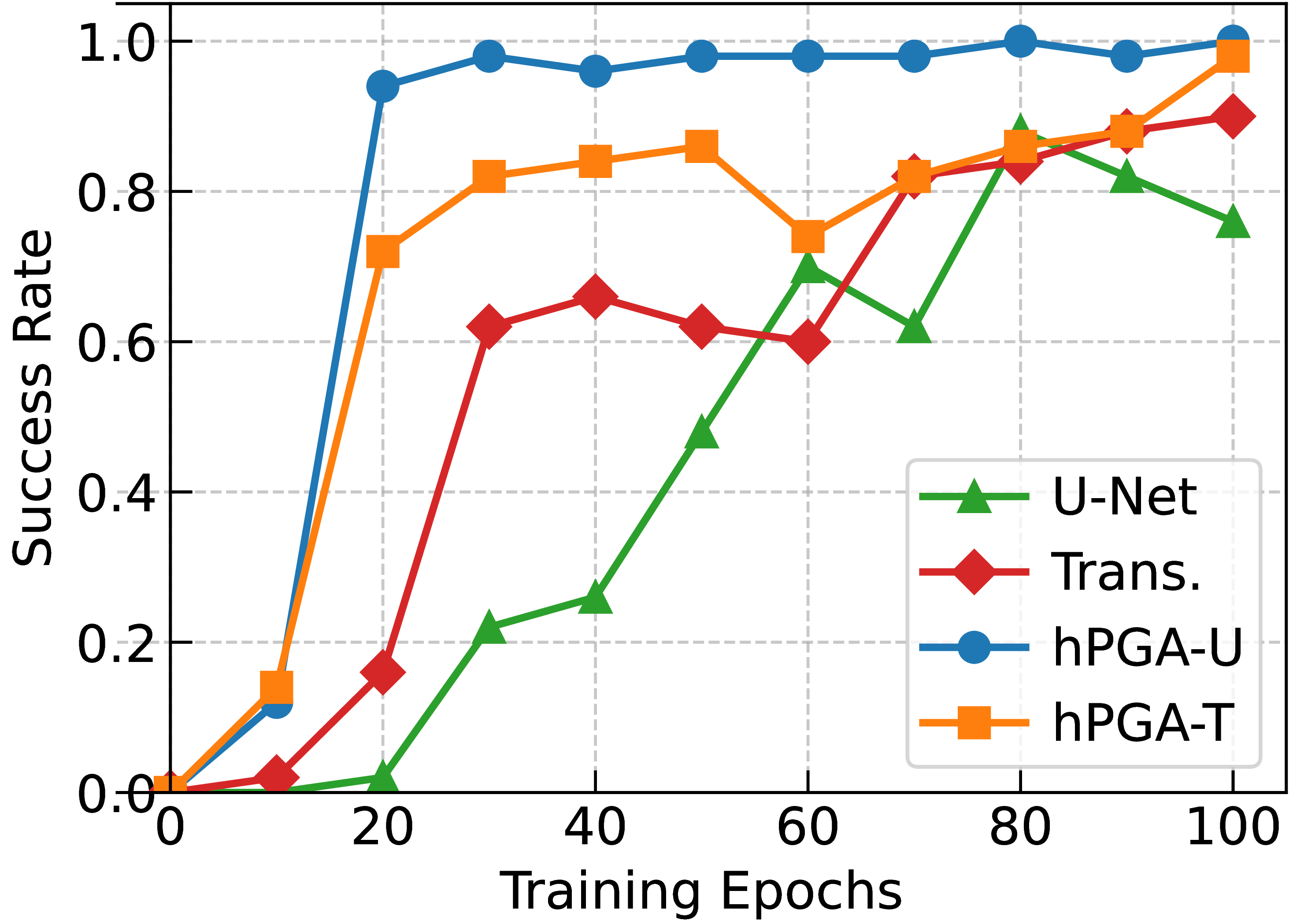}
    \end{minipage}
\end{minipage}
\caption{\textbf{Top}: simulation tasks in robosuite, with colored 3D bounding boxes indicating task-relevant objects. \textbf{Bottom left}: success rates for diffusion policies with different network backbones for various tasks, and mean epoch training time (MET) for each network on all tasks together. \textbf{Bottom right}: plot of success rate for state-based policies with U-Net, Transformer, hPGA-U, and hPGA-T for 100 training epochs of the Stack task.}
\label{fig:sim_exp}
\vspace{-10pt}
\end{figure*}

Although the state encoder and denoising module are jointly trained using the loss $\mathcal{L}_{\text{Encode\&Denoise}}$, the action decoder is intentionally excluded from this objective. This design choice stems from the fact that diffusion models are trained by sampling random denoising steps, where the model predicts noise at arbitrary points in the denoising trajectory in parallel. In contrast, inference proceeds sequentially through all denoising steps in an iterative manner. If the action decoder were used across all training steps, it would become entangled again with the denoising process, requiring it to decode from highly noisy action latents that are not suitable for the geometric inductive bias of P-GATr.

To address this problem, we restrict the decoder's supervision to the final $\eta$ fraction of denoising steps, where $\eta$ is a threshold percentage to mask the loss for decoder. The threshold denoising step is calculated by:
\begin{equation*}
K_{\text{thresh}} = K_{\text{max}} - \left\lfloor \eta \cdot K_{\text{max}} \right\rfloor,
\end{equation*}
where \( K_{\text{max}} \) is the total number of denoising steps and \( \eta \in [0, 1] \) is a tunable parameter. We empirically show that this approach is robust to the choice of \( \eta \) in ablation studies. By limiting supervision to well-denoised action latents, the decoder learns to operate in a structured regime closer to what it encounters during inference, avoiding the need to decode from pure noise.

Since the denoising module \( \epsilon_\theta \) predicts the noise added to the clean latent, we first use its predicted noise \( \hat{\boldsymbol{\varepsilon}} \) to compute the estimated denoised action latent \( \hat{\mathbf{z}}_{\mathbf{a}, 0} \) via the standard reverse process from \citet{ho2020denoising}:
$$
\hat{\mathbf{z}}_{\mathbf{a}, 0} =
\frac{1}{\sqrt{\bar{\alpha}_k}} \left( \mathbf{z}_{\mathbf{a}, k} - \sqrt{1 - \bar{\alpha}_k} \cdot \hat{\boldsymbol{\varepsilon}} \right),
$$
where \( \bar{\alpha}_k \) is the cumulative signal retention factor at timestep \( k \), which we also use for the forward noising process.

The decoder is then trained to reconstruct the ground-truth action multivector sequence \( \mathbf{x}^{t:t+H_p}_{\mathbf{a}} \) from \( \hat{\mathbf{z}}_{\mathbf{a}, 0} \). The decoder loss is defined as:
$$
\mathcal{L}_{\text{Decoder}} =
\mathbf{1}_{\{k \geq K_{\text{thresh}}\}} \cdot
\left\| D_\phi\left( \hat{\mathbf{z}}_{\mathbf{a}, 0} \right) - \mathbf{x}^{t:t+H_p}_{\mathbf{a}} \right\|^2,
$$
where \( D_\phi \) denotes the action decoder, and \( \mathbf{1}_{\{\cdot\}} \) is an indicator function that activates only when the current denoising step \( k \) exceeds the threshold \( K_{\text{thresh}} \).

This staged supervision strategy preserves architectural modularity, leverages geometric inductive biases of PGA, and speeds up training by limiting the decoder's learning signal to geometrically meaningful latents. The overall training objective of hPGA-DP combines both loss components: 
$$\mathcal{L}_{\text{Total}} = \mathcal{L}_{\text{Encode\&Denoise}} + \mathcal{L}_{\text{Decoder}}.$$

\section{Evaluation}
\label{sec:evaluation}

\subsection{Simulation Experiments}

\subsubsection{Experimental Settings}
We evaluate the proposed hPGA-DP framework through simulation experiments and ablation studies across five Robosuite tasks \cite{zhu2020robosuite}, using a 7-DOF Panda Arm. As shown in the top row of Fig.~\ref{fig:sim_exp}, the tasks include: Lift (lift a red cube), Can (sort a can), Stack (stack a red cube on a green one), Square (insert a square nut), and Mug (place a mug into a drawer). Demonstrations for Lift and Can are sourced from Robomimic \cite{robomimic2021}, while those for Stack, Square, and Mug are generated using MimicGen \cite{mandlekar2023mimicgen}. Each dataset contains 200 trajectories, except Mug, which uses 300 due to its higher complexity.

We compare two hPGA-DP variants: hPGA-U and hPGA-T, which use U-Net and Transformer denoising modules, respectively. Given that our focus is on architectural innovation, we compare against U-Net and Transformer as they are the most widely used backbones for diffusion policies, in addition to a standalone P-GATr backbones without the encoder-denoiser-decoder structure. The U-Net and Transformer baselines contain 24M and 35M parameters, which are the same as the respective number of parameters in hPGA-U and hPGA-T. For object pose input, we test both ground-truth state access (State) and vision-based input via 6D pose estimation using PRISM-DP \cite{sun2025prismdp} with FoundationPose \cite{wen2024foundationpose}, leveraging RGB, depth, and generated mesh inputs. 

All models are implemented in PyTorch \cite{NEURIPS2019_9015} and trained on a workstation with an AMD PRO 5975WX CPU, dual NVIDIA RTX 4090 GPUs, and 128GB RAM. Training schedules are: 80 epochs (Lift), 90 (Can), 30 (Stack), 120 (Square), and 100 (Mug). For all hPGA-DP variants, the maximum denoising step is set to $K_{\text{max}} = 100$, and the action decoder loss is applied only during the final $\eta = 0.25$ portion of the denoising steps.

Each policy is evaluated over 50 rollouts, and success rate, which is defined as the fraction of successful trials, is reported as the primary performance metric, while the mean training time per epoch measured in seconds for each network across all tasks is reported as an indication of efficiency.

\subsubsection{Results}
The results in Fig.~\ref{fig:sim_exp} show that hPGA-DP, regardless of using U-Net or Transformer for denoising, consistently outperforms baseline policies that rely solely on these backbones. hPGA-DP achieves strong performance within 100 epochs on most tasks, with the exception of Mug, which involves multi-step interactions and requires more training. Notably, hPGA-U, despite having fewer parameters, often surpasses hPGA-T in performance.

hPGA-DP also demonstrates greater training efficiency. Despite that hPGA-DP takes more time to train per epoch according to the MET column of the bottom left table on Fig. \ref{fig:sim_exp}, it requires many fewer training epochs to converge. For example, in the Stack task (bottom-right plot, Fig.~\ref{fig:sim_exp}), hPGA-DP variants reach high success rates within about 30 epochs, while U-Net-only and Transformer-only baselines require roughly three times more training epochs to match this level.

Policies using P-GATr as the denoising network fail across all tasks due to extremely slow convergence. As corroborated by the concurrent work \cite{zhong2025gagrasp}, training P-GATr for effective denoising requires at least seven days on high-end GPUs, rendering it far less practical than both hPGA-DP and standard diffusion backbones.

\subsubsection{Ablation Studies}
To further analyze the design of hPGA-DP, we conduct two ablation studies: (1) to identify an effective range for the action decoder loss masking threshold $\eta$, and (2) to evaluate whether the performance gains stem primarily from the encoder-denoiser-decoder layout. All experiments are conducted on the Stack task, which converges most rapidly among the benchmarks. Each configuration is evaluated over 5 trials, with 50 rollouts per trial. We report mean and standard deviation of the success rate.

\begin{figure}[t]
    \centering
    \vspace{-12pt}

    \begin{minipage}{\columnwidth}
        \centering
        \includegraphics[width=\linewidth]{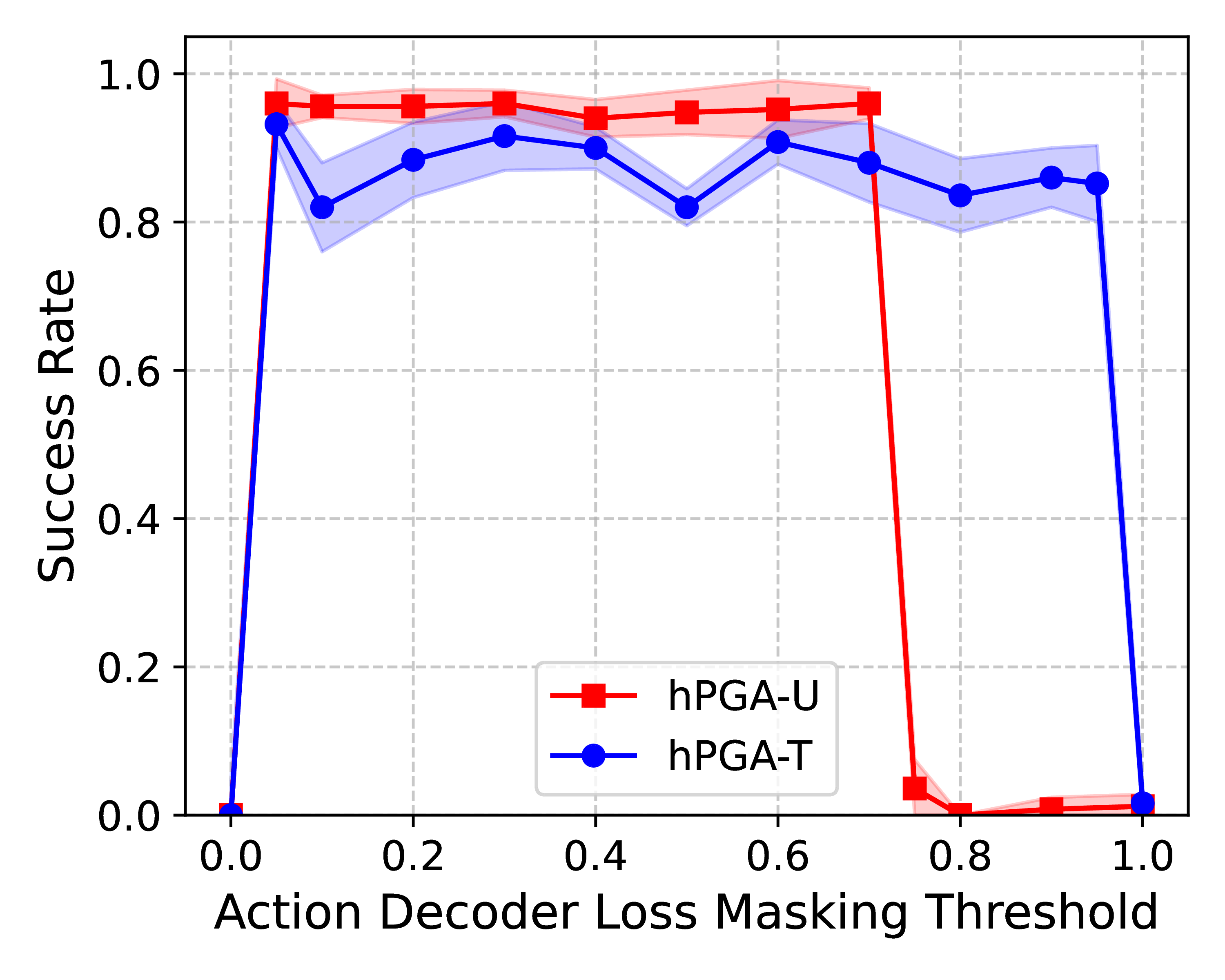}
    \end{minipage}
    
    \vspace{6pt}

    \begin{minipage}{\columnwidth}
        \begin{tabular*}{\linewidth}{@{\extracolsep{\fill}}cccc}
            \toprule
            & \textit{MLP} & \textit{Trans.} & \textit{P-GATr} \\
            \midrule
            \underline{U-Net}  & $0.24{\pm}0.06$ & $0.26{\pm}0.04$ & $\mathbf{0.96{\pm}0.02}$ \\
            \underline{Trans.} & $0.58{\pm}0.04$ & $0.62{\pm}0.04$ & $\mathbf{0.88{\pm}0.05}$ \\
            \bottomrule
        \end{tabular*}
    \end{minipage}

    \caption{\textbf{Top}: Success rate of hPGA-DP under different decoder loss masking thresholds $\eta$, where solid line denotes the mean and shaded region indicates the standard deviation.
    \textbf{Bottom}: Performance of diffusion policies with various combinations of \underline{backbone} (underlined) and \textit{encoder \& decoder} (italicized).}
    \label{fig:ablation}
    \vspace{-15pt}
\end{figure}

For the $\eta$ ablation, we test 10 values from 0.0 to 1.0 in increments of 0.1 for both hPGA-U and hPGA-T, using ground-truth object states. We additionally include two intermediate values in regions showing steep performance drops, resulting in 12 variants per model (top plot, Fig.~\ref{fig:ablation}). Results show that both hPGA-U and hPGA-T are robust to the choice of $\eta$: hPGA-U performs well for $\eta \in [0.05, 0.75]$, while hPGA-T maintains stable performance within $\eta \in [0.05, 0.95]$. The low variance across trials reflects consistency in training.

This robustness aligns with observations by \citet{ho2020denoising}, which suggest that most noise is removed in early denoising steps. Consequently, action latents quickly acquire a coarse geometric structure sufficient for effective decoder training, even if not yet executable. High values of $\eta$ still provide meaningful gradients, while very low $\eta$ limits decoder updates since most samples are excluded.

To evaluate the impact of network layout, we perform a second ablation by varying the encoder and decoder while keeping the denoiser fixed as either Transformer or U-Net. We compare three encoder-decoder types: MLP, Transformer, and P-GATr. The P-GATr variants correspond to hPGA-T and hPGA-U. MLP and Transformer are included due to their widespread use in prior works as encoder and decoder \cite{wu2024deep, loquercio2021learning, dosovitskiy2020image}, while U-Net is typically only used as backbone \cite{sharma2023u, si2024freeu, mitsouras2024u}. For MLP and Transformer variants, we use standard MSE loss \cite{chi2023diffusion}, as they do not operate on multivectors and thus do not require loss masking or $\eta$.

As shown in Fig.\ref{fig:ablation}, using MLP or Transformer as the encoder and decoder does not yield significant improvements over their baselines in Fig.\ref{fig:sim_exp}. This suggests that the gains from hPGA-DP arise not simply from its layout, but from the integration of P-GATr and its geometry-aware training strategy.

\subsection{Real-World Experiments}

\subsubsection{Experimental Settings}
\begin{figure}[t]
    \centering
    \vspace{-12pt}

    \begin{minipage}{\columnwidth}
        \centering
        \includegraphics[width=\linewidth]{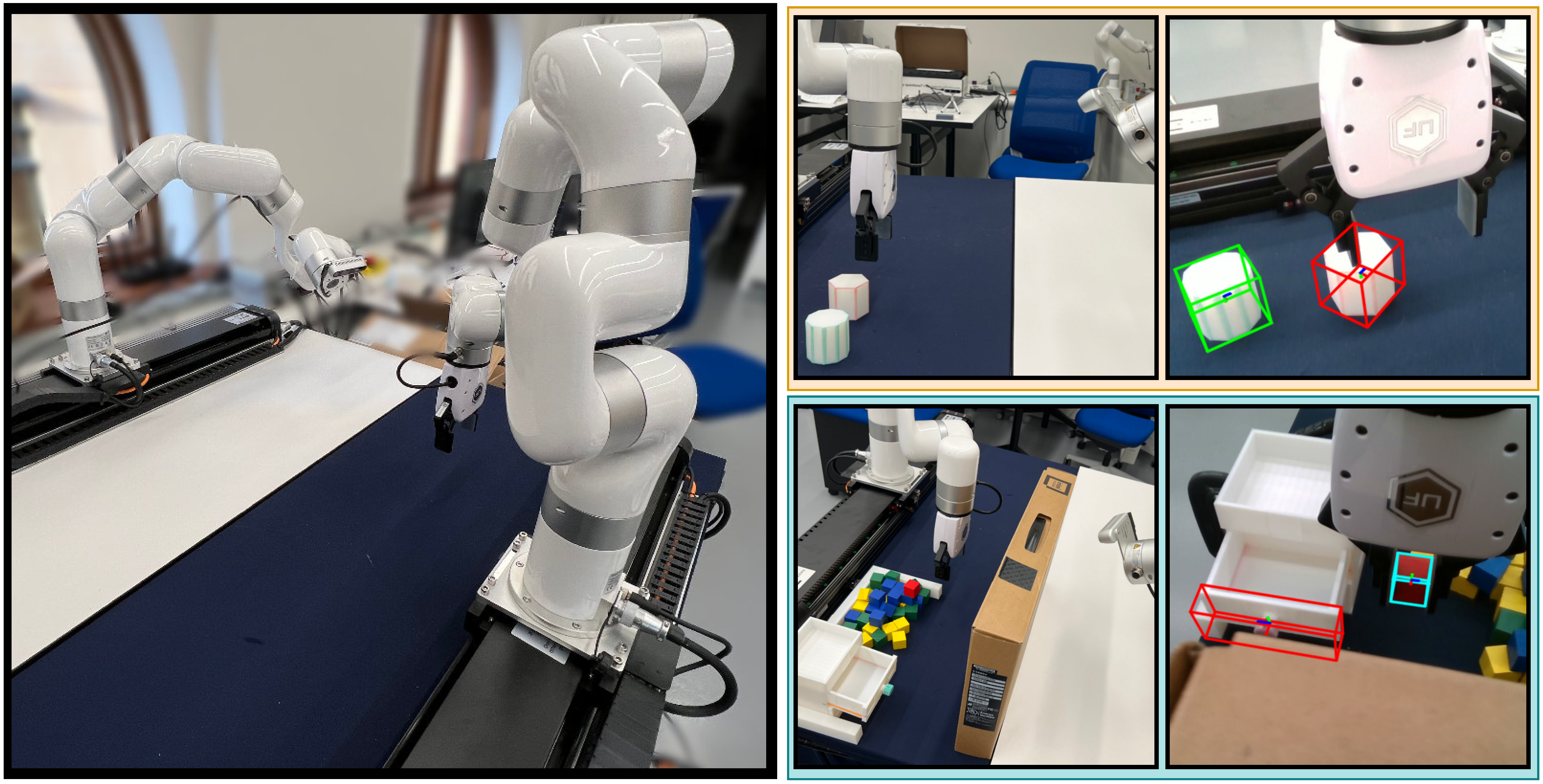}
    \end{minipage}
    
    \vspace{6pt}

    \begin{minipage}{\columnwidth}
        \begin{tabular*}{\linewidth}{l @{\extracolsep{\fill}} cccc}
            \toprule
            \textbf{Network}
            & \multicolumn{2}{c}{\textbf{Block Stack.}}
            & \multicolumn{2}{c}{\textbf{Drawer Inter.}} \\
            & \textbf{SR} & \textbf{CT} & \textbf{SR} & \textbf{CT} \\
            \midrule
            U-Net      & $0.43$ & $\mathbf{9.85}$ & $0.27$ & $\mathbf{14.28}$ \\
            Trans.     & $0.37$ & $12.03$ & $0.40$ & $15.87$ \\
            P-GATr     & $0.00$ & $92.42$ & $0.00$ & $119.63$ \\
            \midrule
            hPGA-U     & $\mathbf{0.97}$ & $16.25$ & $\mathbf{0.90}$ & $22.53$ \\
            hPGA-T     & $0.93$ & $17.69$ & $0.87$ & $26.81$ \\
            \bottomrule
        \end{tabular*}
    \end{minipage}

    \caption{\textbf{Top left}: the dual-arm system for real-world experiments. \textbf{Top right}: top and bottom row show the block stacking task and drawer interaction task respectively. \textbf{Bottom}: results for real-world experiments. SR: success rate, CT: cumulative training time measured in minutes.}
    \label{fig:realworld_exp}
    \vspace{-15pt}
\end{figure}

To further evaluate hPGA-DP, we conduct real-world experiments using a dual-arm setup (top-left image, Fig.~\ref{fig:realworld_exp}) consisting of two 7-DOF xArm7 robots mounted on 1-DOF linear actuators. One arm is equipped with an Intel RealSense D435i depth camera, while the other is fitted with a parallel gripper. The system operates in a look-at end-effector space, where the camera’s orientation is excluded from the state-action space by dynamically adjusting it through inverse kinematics constraints to maintain focus on the gripper \cite{sun2024comparative, rakita2018autonomous}. 

We evaluate our approach on two real-world tasks, visualized in Fig.~\ref{fig:realworld_exp}. The block stacking task (Block Stack.) requires placing a non-cuboid block onto another, while the drawer interaction task (Drawer Inter.) involves clearing a visual occlusion, retrieving a red cube, inserting it into a drawer, and closing the drawer. The blocks and drawer are 3D printed using meshes from \citet{lee2021beyond} and \citet{heo2023furniturebench}, which are also used as ground-truth meshes for evaluation. Each task is trained on a dataset of 200 demonstration rollouts.

The evaluated conditions are the same as the simulation experiment with the exception of those that received ground-truth state information as that is not available in the real-world.  Instead, all conditions inferred object poses using the same 6D pose estimation strategy employed in the simulation experiments---namely, PRISM-DP \citep{sun2025prismdp} with a FoundationPose backend \cite{wen2024foundationpose}.  An example of pose tracking results from a single frame is shown in the top-right panel of Fig.~\ref{fig:realworld_exp}.

Training and inference are performed on the same workstation used for simulation. We report success rate and cumulative training time as performance metrics. Note that training time includes only the forward and backward passes during epoch training, excluding data loading and sampling overhead.

\subsubsection{Results} As shown in the table of Fig.~\ref{fig:realworld_exp}, hPGA-DP achieves significantly higher success rates compared to other network backbones trained for the same number of epochs. While hPGA-DP requires more training time than Transformer and U-Net baselines, those baselines need to be trained for twice as many epochs to match hPGA-DP's performance, which results in 21\% to 36\% higher total training time. These findings are consistent with simulation results, demonstrating that hPGA-DP offers superior performance and training efficiency over traditional diffusion backbones.

\section{Discussion}
\label{sec:discussion}
In this work, we introduced hPGA-DP, a hybrid diffusion policy that successfully integrates PGA to embed strong geometric inductive biases into robot learning.
Our architecture uses P-GATr for state encoding and action decoding alongside a conventional denoising module.
This design, combined with a staged supervision strategy for the decoder, significantly improves task performance and convergence speed over standard baselines by harnessing PGA's geometric reasoning while maintaining practical training efficiency.

While promising, our approach has limitations that suggest avenues for future work.
Although hPGA-DP converges in fewer epochs than traditional diffusion backbones, each epoch takes slightly longer to train. This inefficiency likely stems from our current PyTorch-based implementation, where the default backward pass may poorly handle the intricate blade-wise interactions in PGA. This could be addressed by developing custom compute kernels using lower-level frameworks such as Triton \cite{tillet2019triton} to accelerate the PGA operations.
Solving this challenge would further broaden the applicability of using geometric algebras in robot learning.





\bibliographystyle{plainnat}

\bibliography{refs}

@article{chi2023diffusion,
  title={Diffusion policy: Visuomotor policy learning via action diffusion},
  author={Chi, Cheng and Xu, Zhenjia and Feng, Siyuan and Cousineau, Eric and Du, Yilun and Burchfiel, Benjamin and Tedrake, Russ and Song, Shuran},
  journal={The International Journal of Robotics Research},
  pages={02783649241273668},
  year={2023},
  publisher={SAGE Publications Sage UK: London, England}
}

@article{brehmer2023geometric,
  title={Geometric algebra transformer},
  author={Brehmer, Johann and De Haan, Pim and Behrends, S{\"o}nke and Cohen, Taco S},
  journal={Advances in Neural Information Processing Systems},
  volume={36},
  pages={35472--35496},
  year={2023}
}

@inproceedings{ronneberger2015u,
  title={U-net: Convolutional networks for biomedical image segmentation},
  author={Ronneberger, Olaf and Fischer, Philipp and Brox, Thomas},
  booktitle={Medical image computing and computer-assisted intervention--MICCAI 2015: 18th international conference, Munich, Germany, October 5-9, 2015, proceedings, part III 18},
  pages={234--241},
  year={2015},
  organization={Springer}
}

@article{vaswani2017attention,
  title={Attention is all you need},
  author={Vaswani, Ashish and Shazeer, Noam and Parmar, Niki and Uszkoreit, Jakob and Jones, Llion and Gomez, Aidan N and Kaiser, {\L}ukasz and Polosukhin, Illia},
  journal={Advances in neural information processing systems},
  volume={30},
  year={2017}
}

@article{ho2020denoising,
  title={Denoising diffusion probabilistic models},
  author={Ho, Jonathan and Jain, Ajay and Abbeel, Pieter},
  journal={Advances in neural information processing systems},
  volume={33},
  pages={6840--6851},
  year={2020}
}

@article{podell2023sdxl,
  title={Sdxl: Improving latent diffusion models for high-resolution image synthesis},
  author={Podell, Dustin and English, Zion and Lacey, Kyle and Blattmann, Andreas and Dockhorn, Tim and M{\"u}ller, Jonas and Penna, Joe and Rombach, Robin},
  journal={arXiv preprint arXiv:2307.01952},
  year={2023}
}

@article{guo2023animatediff,
  title={Animatediff: Animate your personalized text-to-image diffusion models without specific tuning},
  author={Guo, Yuwei and Yang, Ceyuan and Rao, Anyi and Liang, Zhengyang and Wang, Yaohui and Qiao, Yu and Agrawala, Maneesh and Lin, Dahua and Dai, Bo},
  journal={arXiv preprint arXiv:2307.04725},
  year={2023}
}

@article{sun2025dynamic,
  title={Dynamic Rank Adjustment in Diffusion Policies for Efficient and Flexible Training},
  author={Sun, Xiatao and Yang, Shuo and Chen, Yinxing and Fan, Francis and Liang, Yiyan and Rakita, Daniel},
  journal={arXiv preprint arXiv:2502.03822},
  year={2025}
}

@article{wang2025hierarchical,
  title={Hierarchical Diffusion Policy: manipulation trajectory generation via contact guidance},
  author={Wang, Dexin and Liu, Chunsheng and Chang, Faliang and Xu, Yichen},
  journal={IEEE Transactions on Robotics},
  year={2025},
  publisher={IEEE}
}

@article{hou2024diffusion,
  title={Diffusion transformer policy},
  author={Hou, Zhi and Zhang, Tianyi and Xiong, Yuwen and Pu, Hengjun and Zhao, Chengyang and Tong, Ronglei and Qiao, Yu and Dai, Jifeng and Chen, Yuntao},
  journal={arXiv preprint arXiv:2410.15959},
  year={2024}
}

@article{wang2025mtdp,
  title={MTDP: Modulated Transformer Diffusion Policy Model},
  author={Wang, Qianhao and Sun, Yinqian and Lu, Enmeng and Zhang, Qian and Zeng, Yi},
  journal={arXiv preprint arXiv:2502.09029},
  year={2025}
}

@article{ze20243d,
  title={3d diffusion policy: Generalizable visuomotor policy learning via simple 3d representations},
  author={Ze, Yanjie and Zhang, Gu and Zhang, Kangning and Hu, Chenyuan and Wang, Muhan and Xu, Huazhe},
  journal={arXiv preprint arXiv:2403.03954},
  year={2024}
}

@article{wang2024equivariant,
  title={Equivariant diffusion policy},
  author={Wang, Dian and Hart, Stephen and Surovik, David and Kelestemur, Tarik and Huang, Haojie and Zhao, Haibo and Yeatman, Mark and Wang, Jiuguang and Walters, Robin and Platt, Robert},
  journal={arXiv preprint arXiv:2407.01812},
  year={2024}
}

@article{reuss2024efficient,
  title={Efficient Diffusion Transformer Policies with Mixture of Expert Denoisers for Multitask Learning},
  author={Reuss, Moritz and Pari, Jyothish and Agrawal, Pulkit and Lioutikov, Rudolf},
  journal={arXiv preprint arXiv:2412.12953},
  year={2024}
}

@article{bilaloglu2024tactile,
  title={Tactile Ergodic Control Using Diffusion and Geometric Algebra},
  author={Bilaloglu, Cem and L{\"o}w, Tobias and Calinon, Sylvain},
  journal={arXiv preprint arXiv:2402.04862},
  year={2024}
}

@article{zhong2025gagrasp,
  title={Gagrasp: Geometric algebra diffusion for dexterous grasping},
  author={Zhong, Tao and Allen-Blanchette, Christine},
  journal={arXiv preprint arXiv:2503.04123},
  year={2025}
}

@book{dorst2009geometric,
  title={Geometric algebra for computer science (revised edition): An object-oriented approach to geometry},
  author={Dorst, Leo and Fontijne, Daniel and Mann, Stephen},
  year={2009},
  publisher={Morgan Kaufmann}
}

@article{dorst2022guided,
  title={A guided tour to the plane-based geometric algebra PGA},
  author={Dorst, Leo and De Keninck, Steven},
  journal={URL https://bivector. net/PGA4CS. html},
  year={2022}
}

@article{low2023geometric,
  title={Geometric algebra for optimal control with applications in manipulation tasks},
  author={L{\"o}w, Tobias and Calinon, Sylvain},
  journal={IEEE Transactions on Robotics},
  volume={39},
  number={5},
  pages={3586--3600},
  year={2023},
  publisher={IEEE}
}

@article{zhu2025conformal,
  title={A Conformal Geometric Algebra Method for Inverse Kinematics Analysis of 6R Robotic Arm},
  author={Zhu, Dongyang and Zhang, Zhonghai and Li, Duanling},
  journal={Journal of Mechanisms and Robotics},
  pages={1--34},
  year={2025}
}

@article{carbajal2024fika,
  title={FIKA: a conformal geometric algebra approach to a fast inverse kinematics algorithm for an anthropomorphic robotic arm},
  author={Carbajal-Espinosa, Oscar and Campos-Mac{\'\i}as, Leobardo and D{\'\i}az-Rodriguez, Miriam},
  journal={Machines},
  volume={12},
  number={1},
  pages={78},
  year={2024},
  publisher={MDPI}
}

@article{sun2023analytical,
  title={An Analytical Method for Sensitivity Analysis of Rigid Multibody System Dynamics Using Projective Geometric Algebra},
  author={Sun, Guangzhen and Ding, Ye},
  journal={Journal of Computational and Nonlinear Dynamics},
  volume={18},
  number={11},
  year={2023},
  publisher={American Society of Mechanical Engineers Digital Collection}
}

@article{sun2023high,
  title={High-order inverse dynamics of serial robots based on projective geometric algebra},
  author={Sun, Guangzhen and Ding, Ye},
  journal={Multibody System Dynamics},
  volume={59},
  number={3},
  pages={337--362},
  year={2023},
  publisher={Springer}
}

@article{ruhe2023clifford,
  title={Clifford group equivariant neural networks},
  author={Ruhe, David and Brandstetter, Johannes and Forr{\'e}, Patrick},
  journal={Advances in Neural Information Processing Systems},
  volume={36},
  pages={62922--62990},
  year={2023}
}

@article{liu2022geometric,
  title={Geometric algebra graph neural network for cross-domain few-shot classification},
  author={Liu, Qifan and Cao, Wenming},
  journal={Applied Intelligence},
  volume={52},
  number={11},
  pages={12422--12435},
  year={2022},
  publisher={Springer}
}

@article{zhong2023geometric,
  title={Geometric algebra-based multiview interaction networks for 3D human motion prediction},
  author={Zhong, Jianqi and Cao, Wenming},
  journal={Pattern Recognition},
  volume={138},
  pages={109427},
  year={2023},
  publisher={Elsevier}
}

@article{roelfs2023graded,
  title={Graded symmetry groups: plane and simple},
  author={Roelfs, Martin and De Keninck, Steven},
  journal={Advances in Applied Clifford Algebras},
  volume={33},
  number={3},
  pages={30},
  year={2023},
  publisher={Springer}
}

@inproceedings{lee2023towards,
  title={Towards foundation models for materials science: The open matsci ml toolkit},
  author={Lee, Kin Long Kelvin and Gonzales, Carmelo and Spellings, Matthew and Galkin, Mikhail and Miret, Santiago and Kumar, Nalini},
  booktitle={Proceedings of the SC'23 Workshops of the International Conference on High Performance Computing, Network, Storage, and Analysis},
  pages={51--59},
  year={2023}
}

@article{brehmer2024lorentz,
  title={A Lorentz-Equivariant Transformer for All of the LHC},
  author={Brehmer, Johann and Bres{\'o}, V{\'\i}ctor and de Haan, Pim and Plehn, Tilman and Qu, Huilin and Spinner, Jonas and Thaler, Jesse},
  journal={arXiv preprint arXiv:2411.00446},
  year={2024}
}

@inproceedings{xiong2020layer,
  title={On layer normalization in the transformer architecture},
  author={Xiong, Ruibin and Yang, Yunchang and He, Di and Zheng, Kai and Zheng, Shuxin and Xing, Chen and Zhang, Huishuai and Lan, Yanyan and Wang, Liwei and Liu, Tieyan},
  booktitle={International conference on machine learning},
  pages={10524--10533},
  year={2020},
  organization={PMLR}
}

@inproceedings{baevskiadaptive,
  title={Adaptive Input Representations for Neural Language Modeling},
  author={Baevski, Alexei and Auli, Michael},
  booktitle={International Conference on Learning Representations}
}

@article{hendrycks2016gaussian,
  title={Gaussian error linear units (gelus)},
  author={Hendrycks, Dan and Gimpel, Kevin},
  journal={arXiv preprint arXiv:1606.08415},
  year={2016}
}

@article{zhu2020robosuite,
  title={robosuite: A modular simulation framework and benchmark for robot learning},
  author={Zhu, Yuke and Wong, Josiah and Mandlekar, Ajay and Mart{\'\i}n-Mart{\'\i}n, Roberto and Joshi, Abhishek and Nasiriany, Soroush and Zhu, Yifeng},
  journal={arXiv preprint arXiv:2009.12293},
  year={2020}
}

@inproceedings{robomimic2021,
  title={What Matters in Learning from Offline Human Demonstrations for Robot Manipulation},
  author={Ajay Mandlekar and Danfei Xu and Josiah Wong and Soroush Nasiriany and Chen Wang and Rohun Kulkarni and Li Fei-Fei and Silvio Savarese and Yuke Zhu and Roberto Mart\'{i}n-Mart\'{i}n},
  booktitle={Conference on Robot Learning (CoRL)},
  year={2021}
}

@article{mandlekar2023mimicgen,
  title={Mimicgen: A data generation system for scalable robot learning using human demonstrations},
  author={Mandlekar, Ajay and Nasiriany, Soroush and Wen, Bowen and Akinola, Iretiayo and Narang, Yashraj and Fan, Linxi and Zhu, Yuke and Fox, Dieter},
  journal={arXiv preprint arXiv:2310.17596},
  year={2023}
}

@article{sun2025prismdp,
  title={PRISM-DP: Spatial Pose-based Observations for Diffusion-Policies via Segmentation, Mesh Generation, and Pose Tracking},
  author={Sun, Xiatao and Chen, Yinxing and Rakita, Daniel},
  journal={arXiv preprint arXiv:2504.20359},
  year={2025}
}

@inproceedings{wen2024foundationpose,
  title={Foundationpose: Unified 6d pose estimation and tracking of novel objects},
  author={Wen, Bowen and Yang, Wei and Kautz, Jan and Birchfield, Stan},
  booktitle={Proceedings of the IEEE/CVF Conference on Computer Vision and Pattern Recognition},
  pages={17868--17879},
  year={2024}
}

@incollection{NEURIPS2019_9015,
  title     = {PyTorch: An Imperative Style, High-Performance Deep Learning Library},
  author    = {Paszke, Adam and Gross, Sam and Massa, Francisco and Lerer, Adam and Bradbury, James and Chanan, Gregory and Killeen, Trevor and Lin, Zeming and Gimelshein, Natalia and Antiga, Luca and Desmaison, Alban and Kopf, Andreas and Yang, Edward and DeVito, Zachary and Raison, Martin and Tejani, Alykhan and Chilamkurthy, Sasank and Steiner, Benoit and Fang, Lu and Bai, Junjie and Chintala, Soumith},
  booktitle = {Advances in Neural Information Processing Systems 32},
  pages     = {8024--8035},
  year      = {2019},
  publisher = {Curran Associates, Inc.},
}

@article{wu2024deep,
  title={Deep learning for optimization of trajectories for quadrotors},
  author={Wu, Yuwei and Sun, Xiatao and Spasojevic, Igor and Kumar, Vijay},
  journal={IEEE Robotics and Automation Letters},
  volume={9},
  number={3},
  pages={2479--2486},
  year={2024},
  publisher={IEEE}
}

@article{loquercio2021learning,
  title={Learning high-speed flight in the wild},
  author={Loquercio, Antonio and Kaufmann, Elia and Ranftl, Ren{\'e} and M{\"u}ller, Matthias and Koltun, Vladlen and Scaramuzza, Davide},
  journal={Science Robotics},
  volume={6},
  number={59},
  pages={eabg5810},
  year={2021},
  publisher={American Association for the Advancement of Science}
}

@article{dosovitskiy2020image,
  title={An image is worth 16x16 words: Transformers for image recognition at scale},
  author={Dosovitskiy, Alexey and Beyer, Lucas and Kolesnikov, Alexander and Weissenborn, Dirk and Zhai, Xiaohua and Unterthiner, Thomas and Dehghani, Mostafa and Minderer, Matthias and Heigold, Georg and Gelly, Sylvain and others},
  journal={arXiv preprint arXiv:2010.11929},
  year={2020}
}

@article{sharma2023u,
  title={U-Net model with transfer learning model as a backbone for segmentation of gastrointestinal tract},
  author={Sharma, Neha and Gupta, Sheifali and Koundal, Deepika and Alyami, Sultan and Alshahrani, Hani and Asiri, Yousef and Shaikh, Asadullah},
  journal={Bioengineering},
  volume={10},
  number={1},
  pages={119},
  year={2023},
  publisher={MDPI}
}

@inproceedings{si2024freeu,
  title={Freeu: Free lunch in diffusion u-net},
  author={Si, Chenyang and Huang, Ziqi and Jiang, Yuming and Liu, Ziwei},
  booktitle={Proceedings of the IEEE/CVF Conference on Computer Vision and Pattern Recognition},
  pages={4733--4743},
  year={2024}
}

@article{mitsouras2024u,
  title={U-Sketch: An Efficient Approach for Sketch to Image Diffusion Models},
  author={Mitsouras, Ilias and Tsonis, Eleftherios and Tzouveli, Paraskevi and Voulodimos, Athanasios},
  journal={arXiv preprint arXiv:2403.18425},
  year={2024}
}

@article{sun2024comparative,
  title={A Comparative Study on State-Action Spaces for Learning Viewpoint Selection and Manipulation with Diffusion Policy},
  author={Sun, Xiatao and Fan, Francis and Chen, Yinxing and Rakita, Daniel},
  journal={arXiv preprint arXiv:2409.14615},
  year={2024}
}

@inproceedings{rakita2018autonomous,
  title={An autonomous dynamic camera method for effective remote teleoperation},
  author={Rakita, Daniel and Mutlu, Bilge and Gleicher, Michael},
  booktitle={Proceedings of the 2018 ACM/IEEE International Conference on Human-Robot Interaction},
  pages={325--333},
  year={2018}
}

@inproceedings{tillet2019triton,
  title={Triton: an intermediate language and compiler for tiled neural network computations},
  author={Tillet, Philippe and Kung, Hsiang-Tsung and Cox, David},
  booktitle={Proceedings of the 3rd ACM SIGPLAN International Workshop on Machine Learning and Programming Languages},
  pages={10--19},
  year={2019}
}

@inproceedings{lee2021beyond,
  title={Beyond pick-and-place: Tackling robotic stacking of diverse shapes},
  author={Lee, Alex X and Devin, Coline Manon and Zhou, Yuxiang and Lampe, Thomas and Bousmalis, Konstantinos and Springenberg, Jost Tobias and Byravan, Arunkumar and Abdolmaleki, Abbas and Gileadi, Nimrod and Khosid, David and others},
  booktitle={5th Annual Conference on Robot Learning},
  year={2021}
}

@article{heo2023furniturebench,
  title={Furniturebench: Reproducible real-world benchmark for long-horizon complex manipulation},
  author={Heo, Minho and Lee, Youngwoon and Lee, Doohyun and Lim, Joseph J},
  journal={The International Journal of Robotics Research},
  pages={02783649241304789},
  year={2023},
  publisher={SAGE Publications Sage UK: London, England}
}

@article{janner2022planning,
  title={Planning with diffusion for flexible behavior synthesis},
  author={Janner, Michael and Du, Yilun and Tenenbaum, Joshua B and Levine, Sergey},
  journal={arXiv preprint arXiv:2205.09991},
  year={2022}
}


\end{document}